\documentclass[journal]{IEEEtran}
\usepackage[english]{babel}

\usepackage[linesnumbered,ruled]{algorithm2e}
\usepackage{algpseudocode}
\usepackage{color}
\definecolor{g}{gray}{0.6}

\usepackage{amsthm}

\usepackage{multirow}
\usepackage{subcaption}
\usepackage{graphicx}
\usepackage{longtable}

\usepackage{soul}

\usepackage{soul}
\usepackage{booktabs}
\usepackage{arydshln}
\usepackage{amsmath}
\usepackage{amssymb}
\usepackage{float}
\usepackage{supertabular}
\usepackage{arydshln}

\usepackage{hyperref}
\usepackage{subcaption}
\usepackage{multicol}
\usepackage{graphicx}

\ifCLASSINFOpdf
\else
\fi
\begin{document}
%
\title{Inverting The Generator Of A Generative Adversarial Network}


\author{Antonia~Creswell and Anil~A~Bharath, Imperial College London
\thanks{e-mail: ac2211@ic.ac.uk.}}


%


\maketitle

\begin{abstract}


Generative adversarial networks (GANs) learn a deep generative model that is able to synthesise novel, high-dimensional data samples. New data samples are synthesised by passing latent samples, drawn from a chosen prior distribution, through the generative model. Once trained, the latent space exhibits interesting properties, that may be useful for down stream tasks such as classification or retrieval. Unfortunately, GANs do not offer an ``inverse model'', a mapping from data space back to latent space, making it difficult to infer a latent representation for a given data sample. In this paper, we introduce a technique, \textit{inversion}, to project data samples, specifically images, to the latent space using a pre-trained GAN. Using our proposed \textit{inversion} technique, we are able to identify which attributes of a dataset a trained GAN is able to model and quantify GAN performance, based on a reconstruction loss. We demonstrate how our proposed \textit{inversion} technique may be used to quantitatively compare performance of various GAN models trained on three image datasets. We provide code for all of our experiments\footnote{\url{https://github.com/ToniCreswell/InvertingGAN}}.



\end{abstract}




%
\IEEEpeerreviewmaketitle

\section{Introduction}

Generative adversarial networks (GANs) \cite{radford2015unsupervised, goodfellow2014generative} are a class of generative model which are able to synthesise novel, realistic looking images of faces, digits and street numbers \cite{radford2015unsupervised}. GANs involve two networks: a generator, $G$, and a discriminator, $D$. The generator, $G$, is trained to generate synthetic images, taking a random vector, $z$, drawn from a prior distribution, $P(Z)$, as input. The prior is often chosen to be a normal or uniform distribution.

Radford et al. \cite{radford2015unsupervised} demonstrated that generative adversarial networks (GANs) learn a ``rich linear structure", meaning that algebraic operations in $Z$-space often lead to semantically meaningful synthetic samples in image space. Since images represented in $Z$-space are often meaningful, direct access to a $z \in Z$ for a given image, $x \in X$ may be useful for discriminative tasks such as retrieval or classification. Recently, it has also become desirable to be able to access $Z$-space in order to manipulate original images \cite{zhu2016generative}. Thus, there are many reasons we may wish to invert the generator.

Typically, inversion is achieved by finding a vector $z \in Z$ which when passed through the generator produces an image that is very similar to the target image. If no suitable $z$ exists, this may be an indicator that the generator is unable to model either the whole image or certain attributes of the image. We give a concrete example in Section \ref{rec:shoes_rec}. Therefore, inverting the generator, additionally, provides interesting insights to highlight what a trained GAN has learned.


Mapping an image, from image space, $X$, to $Z$-space is non-trivial, as it requires inversion of the generator, which is often a many layered, non-linear model \cite{radford2015unsupervised, goodfellow2014generative, chen2016infogan}. Dumoulin et al. \cite{dumoulin2016adversarially} (ALI) and Donahue et al. (BiGAN) \cite{donahue2016adversarial} proposed learning a third, decoder network alongside the generator and discriminator to map image samples back to $Z$-space. Collectively, they demonstrated results on MNIST, ImageNet, CIFAR-10 and SVHN and CelebA. However, reconstructions of inversions are often poor. Specifically, reconstructions of inverted MNIST digits using methods of Donahue et al. \cite{donahue2015long}, often fail to preserve style and character class. Recently, Li et al. \cite{li2017alice} proposed method to improve reconstructions, however drawbacks to these approaches \cite{li2017alice, donahue2016adversarial, dumoulin2016adversarially} include the need to train a third network which increases the number of parameters that have to be learned, increasing the chances of over-fitting. The need to train an extra network, along side the GAN, also means that inversion cannot be performed on pre-trained networks.

A more serious concern, when employing a decoder model to perform inversion, is that its value as a diagnostic tool for evaluating GANs is hindered. GANs suffer from several pathologies including over-fitting, that we may be able to detect using inversion. If an additional encoder model is trained to perform inversion \cite{li2017alice, donahue2016adversarial, dumoulin2016adversarially,luo2017learning}, the encoder itself may over-fit, thus not portraying the true nature of a trained GAN. Since our approach does not involve training an additional encoder model, we may use our approach for ``trouble shooting'' and evaluating different pre-trained GAN models. 





In this paper, we make the following contributions:
\begin{itemize}
    \item We propose a novel approach to invert the generator of any pre-trained GAN, provided that the computational graph for the generator network is available (Section \ref{sec:method}).
    \item We demonstrate that, we are able to infer a $Z$-space representation for a target image, such that when passed through the GAN, it produces a sample visually similar to the target image (Section \ref{sec:results}).
    \item We demonstrate several ways in which our proposed inversion technique may be used to \textbf{both qualitatively}  (Section \ref{rec:shoes_rec})  \textbf{and quantitatively} compare GAN models (Section \ref{sec:troubleshooting}).
    \item Additionally, we show that batches of $z$ samples can be inferred from batches of image samples, which improves the efficiency of the inversion process by allowing multiple images to be inverted in parallel (Section \ref{sec:batch}).
\end{itemize}

We begin, by describing our proposed inversion technique.



\section{Method: Inverting The Generator}
\label{sec:method}

For a target image, $x \in \Re^{m \times m}$ we want to infer the $Z$-space representation, $z \in Z$, which when passed through the trained generator produces an image very similar to $x$. We refer to the process of inferring $z$ from $x$ as \textit{inversion}. This can be formulated as a minimisation problem:

\begin{equation} \label{cost}
    z^* = \min_z   {- \mathbb E}_x \log[G(z)] 
\end{equation}

Provided that the computational graph for $G(z)$ is known, $z^*$ can be calculated via gradient descent methods, taking the gradient of $G$ w.r.t. $z$. This is detailed in Algorithm \ref{min}.


  
    

\begin{algorithm}

\caption{Algorithm for inferring $z^* \in \Re^d$, the latent representation for an image $x \in \Re^{m \times m}$.}\label{min}
\KwResult{Infer($x$)}
$z^* \sim P_z(Z)$ \;
\While{NOT converged}
{
    $L \gets -( x \log [G(z^*)] + (1-x) \log [1-G(z^*)])$\;
    $z^*\gets z^* - \alpha \nabla_z L $\;
}
\textbf{return} $z^*$ \;

\end{algorithm}

Provided that the generator is deterministic, each $z$ value maps to a single image, $x$. A single $z$ value cannot map to multiple images. However, it is possible that a single $x$ value may map to several $z$ representations, particularly if the generator has collapsed \cite{salimans2016improved}. This suggests that there may be multiple possible $z$ values to describe a single image. This is very different to a discriminative model, where multiple images, may often be described by the same representation vector \cite{mahendran2015understanding}, particularly when a discriminative model learns representations tolerant to variations.

The approach described in Algorithm \ref{min} is similar in spirit to that of Mahendran et al. \cite{mahendran2015understanding}, however instead of inverting a representation to obtain the image that was responsible for it, we invert an image to discover the latent representation that generated it. 




\subsection{Inverting A Batch Of Samples}
\label{sec:batch}


Algorithm \ref{min} shows how we can invert a single data sample, however it may not be efficient to invert single images at a time, rather, a more practical approach is to invert many images at once. We will now show that we are able to invert batches of examples.

Let \textbf{z}$_b \in \Re^{B \times n}$, \textbf{z}$_b=\{z_1, z_2, ... z_B\}$ be a batch of $B$ samples of $z$. This will map to a batch of image samples \textbf{x}$_b \in \Re^{B \times m \times m}$, \textbf{x}$_b=\{x_1, x_2, ... x_B\}$. For each pair $(z_i, x_i)$, $i \in \{1...B\}$, a loss $L_i$, may be calculated. The update for $z_i$ would then be $z_i \gets z_i - \alpha \frac{d L_i}{d z_i}$ 

If reconstruction loss is calculated over a batch, then the batch reconstruction loss would be $\sum_{i=\{1,2...B\}} L_i$, and the update would be: 
\begin{equation}
    \nabla_{\textbf{z}_b} L = \frac{\partial \sum_{i \in \{1,2,...B\}} L_i}{\partial (\textbf{z}_b)}
\end{equation}
\begin{equation}
    =\frac{\partial(L_1+L_2...+L_i)}{\partial(\textbf{z}_b)}
\end{equation}
\begin{equation}
    = \frac{dL_1}{dz_1}, \frac{dL_2}{dz_2}, ... \frac{dL_B}{dz_B}
\end{equation}

Each reconstruction loss depends only on $G(z_i)$, so $L_i$ depends only on $z_i$, which means $\frac{\partial L_i}{\partial z_j}=0$, for all $i\not=j$. This shows that $z_i$ is updated only by reconstruction loss $L_i$, and the other losses do not contribute to the update of $z_i$, meaning that it is valid to perform updates on batches.


\subsection{Using Prior Knowledge Of P(Z)}

A GAN is trained to generate samples from a $z \in Z$ where the distribution over $Z$ is a chosen prior distribution, $P(Z)$. $P(Z)$ is often a multivariate Gaussian or uniform distribution. If $P(Z)$ is a multivariate uniform distribution, $\mathcal{U}[a,b]$, then after updating $z^*$, it can be clipped to be between $[a,b]$. This ensures that $z^*$ lies in the probable regions of $Z$. If $P(Z)$ is a multivariate Gaussian Distribution, $\mathcal{N}[\mathbf{\mu},\mathbf{\sigma}^2]$, regularisation terms may be added to the cost function, penalising samples that have statistics that are not consistent with $P(Z)=\mathcal{N}[\mathbf{\mu},\mathbf{\sigma}^2]$.

If $z \in Z$ is a vector of length $d$ and each of the $d$ elements in $z \in \Re^d$ are drawn independently and from identical distributions, we may be able to add a regularisation term to the loss function. For example, if $P(Z)$ is a multivariate Gaussian distribution, then elements in a single $z$ are independent and identically drawn from a Gaussian distribution. Therefore we may calculate the likelihood of an encoding, z, under a multivariate Gaussian distribution by evaluating:

\[ \log P(z) = \log P(z^1, ..., z^d) = \frac{1}{d}\sum_{i=0}^d \log \mathcal{P}(z^i) \] 
where $z^i$ is the $i^{th}$ element in a latent vector z and $\mathcal{P}$ is the probability density function of a (univariate) Gaussian, which may be calculated analytically. Our new loss function may be given by:



\begin{equation} \label{eqn:regLoss}
     L(z,x) = {\mathbb E}_x \log[G(z)] - \beta \log P(z)
\end{equation}

by minimising this loss function (Equation \ref{eqn:regLoss}), we encourage $z^*$ to come from the same distribution as the prior.




\section{Relation to Previous Work}

In this paper, we build on our own work\footnote{Creswell, Antonia, and Anil Anthony Bharath. "Inverting The Generator Of A Generative Adversarial Network." arXiv preprint arXiv:1611.05644 (2016). This paper was previously accepted at the NIPS Workshop on Adversarial Training, which was made available as a non archival pre-print only}. We have augmented the paper, by performing additional experiments on a shoe dataset \cite{luo2017learning} and CelebA, as well as repeating experiments on the Omniglot dataset using the DCGAN model proposed by Radford et al. \cite{radford2015unsupervised} rather than our own network \cite{creswell2016task}. In addition to proposing a novel approach for mapping data samples to their corresponding latent representation, we show how our approach may be used to quantitatively and qualitatively compare models.


Our approach to inferring $z$ from $x$ is similar to the previous work of Zhu et al. \cite{zhu2016generative}, however we make several additional contributions.

Specifically, Zhu et al. \cite{zhu2016generative} calculate reconstruction loss, by comparing the features of $x$ and $G(z^*)$ extracted from layers of AlexNet, a CNN trained on natural scenes. This approach is likely to fail if generated samples are not of natural scenes (e.g. Omniglot handwritten characters). Our approach considers pixel-wise loss, providing an approach that is generic to the dataset. Further, if our intention is to use the inversion to better understand the GAN model, it is essential not to incorporate information from other pre-trained networks in the inversion process.


An alternative class of inversion methods involves training a separate encoding network to learn a mapping from image samples, $x$ to latent samples $z$. Li et al \cite{li2017alice}, Donahue et al. \cite{donahue2016adversarial} and Dumoulin et al. \cite{dumoulin2016adversarially} propose learning the encoder along side the GAN. Unfortunately, training an additional encoder network may encourage over-fitting, resulting in poor image reconstruction. Further, this approach may not be applied to pre-trained models. 

On the other hand, Luo et al.\footnote{Luo et al. cite our pre-print} \cite{luo2017learning}, train an encoding network after the GAN has been trained, which means that their approach may be applied to pre-trained models. One concern about the approach of Luo et al. \cite{luo2017learning}, is that it may not be an accurate reflection of what the GAN has learned, since the learned decoder may over-fit to the examples it is trained on. For this reason, the approach of Luo et al. \cite{luo2017learning} may not be suitable for inverting image samples that come from a different distribution to the training data. Evidence of over-fitting may be found in Luo et al. \cite{luo2017learning}-Figure 2, where ``original'' image samples being inverted are synthesised samples, rather than samples from a test set data; in other words Luo et al. \cite{luo2017learning} show results (Figure 2) for inverting synthesised samples, rather than \textit{real} image samples from a test set. 

In contrast to Luo et al. \cite{luo2017learning}, we demonstrate our inversion approach on data samples drawn from test sets of real data samples. To make inversion more challenging, in the case of the Omniglot dataset, we invert image samples that come from a different distribution to the training data. We invert image samples from the Omniglot handwritten characters dataset that come from a different set of alphabets to the set used to train the (Omniglot) GAN  (Figure \ref{fig:omni_rec}). Our results will show that, using our approach, we are still able to recover a latent encoding that captures \textbf{most} features of the test data samples.


Finally, previous inversion approaches that use learned encoder models \cite{li2017alice, donahue2016adversarial, dumoulin2016adversarially, luo2017learning} may not be suitable for ``trouble shooting", as symptoms of the GAN may be exaggerated by an encoder that over-fits. We discuss this in more detail in Section \ref{sec:troubleshooting}.









\section{``Pre-trained'' Models}
In this section we discuss the training and architecture details of several different GAN models, trained on $3$ different datasets, which we will use for our inversion experiments detailed in Section \ref{sec:experiments}. We show results on total of $10$ trained models (Sections \ref{sec:results} and \ref{sec:troubleshooting}).

\subsection{Omniglot}
\begin{figure}
\centering 
    \begin{subfigure}{\columnwidth}
    \includegraphics[width=0.9\columnwidth]{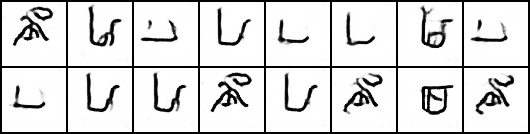} 
    \caption{\textbf{Synthetic Omniglot Samples from a GAN}.}
    \end{subfigure}
    \begin{subfigure}{\columnwidth}
    \includegraphics[width=0.9\columnwidth]{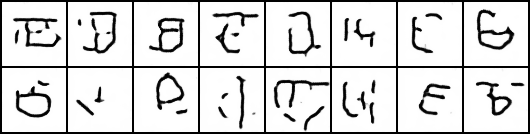}
    \caption{\textbf{Synthetic Omniglot Samples from a WGAN}.} 
    \end{subfigure}
\centering
\caption{\textbf{Synthetic Omniglot samples:} Shows samples synthesised using a (a) GAN and (b) WGAN.}
\label{fig:omni_gen}
\end{figure}

The Omniglot dataset \cite{lake2015human} consists of characters from $50$ different alphabets, where each alphabet has at least $14$ different characters. The Omniglot dataset has a background dataset, used for training and a test dataset. The background set consists of characters from $30$ writing systems, while the test dataset consists of characters from the other $20$. Note, characters in the training and testing dataset come from different writing systems. We train both a DCGAN \cite{radford2015unsupervised} and a WGAN \cite{arjovsky2017wasserstein} using a latent representation of dimension, $d=100$. The WGAN \cite{arjovsky2017wasserstein} is a variant of the GAN that is easier to train and less likely to suffer from mode collapse; mode collapse is where synthesised samples look similar to each other. All GANs are trained with additive noise whose standard deviation decays during training \cite{arjovsky2017towards}.  Figure \ref{fig:omni_gen} shows Omniglot samples synthesised using the trained models. Though it is clear from Figure \ref{fig:omni_gen}(a), that the GAN has collapsed, because the generator is synthesising similar samples for different latent codes, it is less clear to what extent the WGAN  (Figure \ref{fig:omni_gen}(b)) may have collapsed or over-fit. It is also unclear from Figure \ref{fig:omni_gen}(b) what representative power, the (latent space of the)  WGAN has. Results in sections \ref{sec:results} and \ref{sec:troubleshooting} will provide more insight into the representations learned by these models.


\subsection{Shoes}
The shoes dataset \cite{yu2014fine} consists of c.$50,000$ examples of shoes in RGB colour, from $4$ different categories and over $3000$ different subcategories. The images are of dimensions $128\times128$. We leave $1000$ samples out for testing and use the rest for training. We train two GANs using the DCGAN \cite{radford2015unsupervised} architecture. We train one DCGAN with full sized images and the second we train on $64\times64$ images. The networks were trained according to the setup described by Radford et al. \cite{radford2015unsupervised}, using a multivariate Gaussian prior. We also train a WGAN \cite{arjovsky2017wasserstein} on full sized images. All GANs are trained with additive noise whose standard deviation decays during training \cite{arjovsky2017towards}. Figure \ref{fig:shoe_gen} shows samples randomly synthesised using the DCGAN models trained on shoes. The samples look quite realistic, but again, they do not tell us much about the representations learned by the GANs.

\begin{figure}
\begin{subfigure}{\columnwidth}
    \centering
    \includegraphics[width=\columnwidth]{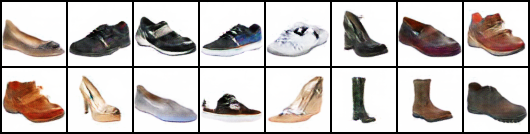}
    \caption{\textbf{Synthetic Shoe Samples from a DCGAN $64\times64$}.}
    \label{fig:my_label}
\end{subfigure}
\begin{subfigure}{\columnwidth}
    \centering
    \includegraphics[width=\columnwidth]{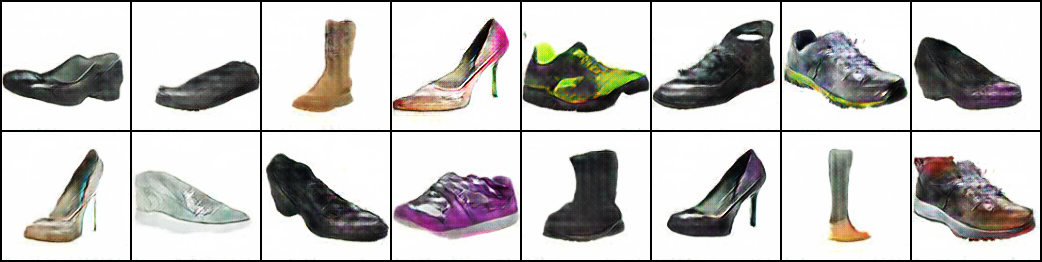}
    \caption{\textbf{Synthetic Shoe Samples from a DCGAN $128\times128$}.}
    \label{fig:my_label}
\end{subfigure}
\begin{subfigure}{\columnwidth}
    \centering
    \includegraphics[width=\columnwidth]{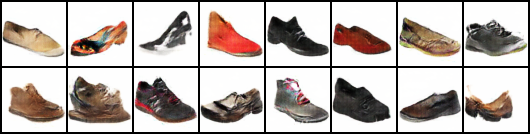} 
    \caption{\textbf{Synthetic Shoe Samples from a WGAN $64\times64$}.}
    \label{fig:my_label}
\end{subfigure}
\caption{\textbf{Shoe samples synthesised using GANs:} Shows samples from DCGANs trained on  (a) lower resolution ($64\times64$) images, (b) higher resolution images ($128\times128$) and (c) samples from a WGAN. }
\label{fig:shoe_gen}
\end{figure}

\subsection{CelebA}
\begin{figure}
\begin{subfigure}{\columnwidth}
    \centering
    \includegraphics[width=\columnwidth]{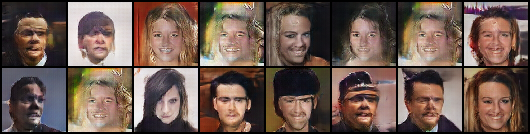} 
    \caption{\textbf{Synthetic Face Samples from GAN trained with noise.}}
    \label{fig:my_label}
\end{subfigure}
\begin{subfigure}{\columnwidth}
    \centering
    \includegraphics[width=\columnwidth]{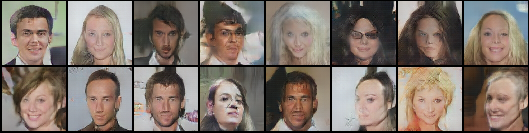} 
    \caption{\textbf{Synthetic Face Samples from a GAN trained without noise.}}
    \label{fig:my_label}
\end{subfigure}
\begin{subfigure}{\columnwidth}
    \centering
    \includegraphics[width=\columnwidth]{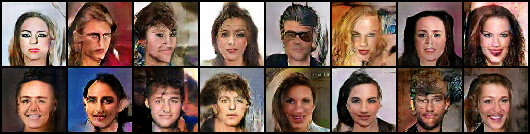} 
    \caption{\textbf{Synthetic Face Samples from a WGAN trained with noise.}}
    \label{fig:face_gen}
\end{subfigure}
\caption{\textbf{Celebrity faces synthesised using GANs:} Shows samples from DCGANs trained (a) without noise and (b) with noise, and (c) samples from a WGAN.}
\label{fig:shoe_gen}
\end{figure}

The CelebA dataset consists of $250,000$ celebrity faces, in RGB colour. The images are of dimensions $64\times64$ pixels. We leave $1000$ samples out for testing and use the rest for training. We train three models, a DCGAN and WGAN trained with decaying noise \cite{arjovsky2017towards} and a DCGAN trained without noise. The networks are trained according to the setup described by Radford et al. \cite{radford2015unsupervised}. Figure \ref{fig:face_gen} shows examples of faces synthesised with and without noise. It is clear from Figure \ref{fig:face_gen}(a) that the GAN trained without noise has collapsed, synthesising similar examples for different latent codes. The WGAN produces the sharpest and most varied samples. However, these samples do not provide sufficient information about the representation power of the models.

\section{Experiments}
\label{sec:experiments}
To obtain latent representations, $z^*$ for a given image $x$ we apply our proposed inversion technique to a batch of randomly selected test images, $x \in X$. To invert a batch of image samples, we minimised the cost function described by Equation (\ref{eqn:regLoss}). In most of our experiments we use $\beta=0.01$, unless stated otherwise, and update candidate $z^*$ using an RMSprop optimiser, with a learning of rate $0.01$.

A valid inversion process should map a target image sample, $x\in X$ to a $z^* \in Z$, such that when $z^*$ is passed through the generative part of the GAN, it produces an image, $G(z^*)$, that is close to the target image, $x$. However, the quality of the reconstruction, depends heavily on the latent representation that the generative model has learned. In the case where a generative model is only able to represent some attributes of the target image, $x$, the reconstruction, $G(z^*)$ may only partially reconstruct $x$.

Thus, the purpose of our experiments is two fold:
\begin{enumerate}
    \item To demonstrate qualitatively, through reconstruction, ($G(z^*)$), that for most well trained GANs, our inversion process is able to recover a latent code, $z^*$, that captures \textbf{most of the important features} of a target image (Section \ref{sec:results}).
    \item To demonstrate how our proposed inversion technique may be used to both qualitatively (Section \ref{rec:shoes_rec}). and \textbf{quantitatively} compare GAN models (Section \ref{sec:troubleshooting}).
\end{enumerate}

\section{Reconstruction Results}
\label{sec:results}

\subsection{Omniglot}
\begin{figure}[h]
\centering
    \begin{subfigure}{\columnwidth}
    \centering
        \includegraphics[width=\columnwidth]{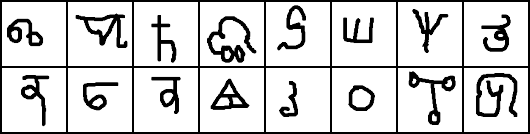}
        \caption{ Target Omniglot handwritten characters, $x$, from alphabets different to those seen during training.}
    \end{subfigure}
    \begin{subfigure}{\columnwidth}
        \includegraphics[width=\columnwidth]{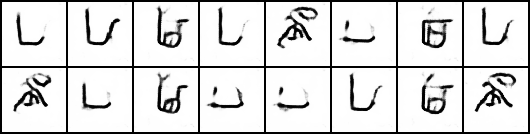} 
        \caption{Reconstructed data samples, $G(z^*)$, using a GAN.}
    \end{subfigure}
    \begin{subfigure}{\columnwidth}
        \includegraphics[width=\columnwidth]{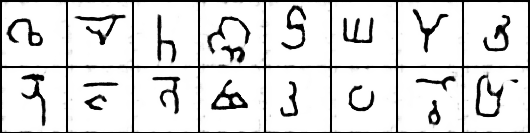} 
        \caption{Reconstructed data samples, $G(z^*)$, using a WGAN.}
    \end{subfigure}
    \begin{subfigure}{\columnwidth}
        \includegraphics[width=\columnwidth]{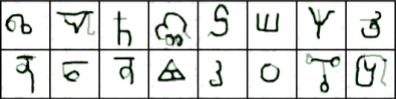}
        \caption{Reconstructed data samples, $G(z^*)$, using a WGAN overlaid with $x$.}
    \end{subfigure}
    \caption{\textbf{Reconstruction of Omniglot handwritten characters.}}
    \label{fig:omni_rec}
\end{figure}

The Omniglot inversions are particularly challenging, as we are trying to find a set of $z^*$'s for a set of characters, $x$, from alphabets that were not in the training data. The inversion process will involve finding representations for data samples from alphabets that it has not seen before, using information about alphabets that it has seen. The original and reconstructed samples are shown in Figure \ref{fig:omni_rec}.


In our previous work,\footnote{https://arxiv.org/abs/1611.05644} we showed that given the ``correct'' architecture, we are able to find latent representations that lead to excellent reconstructions. However, here we focus on evaluating standard models \cite{radford2015unsupervised} and we are particularly interested in detecting (and quantifying) where models fail, especially since visual inspection of synthesised samples may not be sufficient to detect model failure. 


It is clear from Figure \ref{fig:omni_gen} that the GAN has over-fit, however it was less clear whether or not the WGAN has, since samples appeared to be more varied. By attempting to perform inversion, we can see that the WGAN has indeed over-fit, as it is only able to partially reconstruct the target data samples. In the next section (Section \ref{sec:troubleshooting}), we quantitatively compare the extent to which the GAN and WGAN trained on the Omniglot dataset have over-fit.


\subsection{Shoes}
\label{rec:shoes_rec}
\begin{figure}[h]
    \begin{subfigure}{\columnwidth}
        \includegraphics[width=\columnwidth]{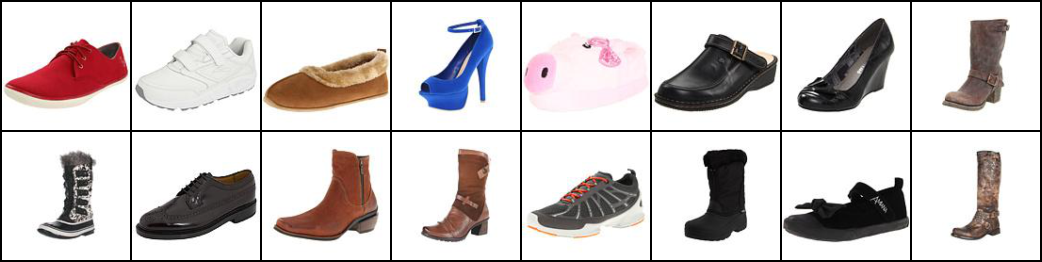}
        \caption{Shoe data samples, $x$, from a test set}
    \end{subfigure}
    \begin{subfigure}{\columnwidth}
        \includegraphics[width=\columnwidth]{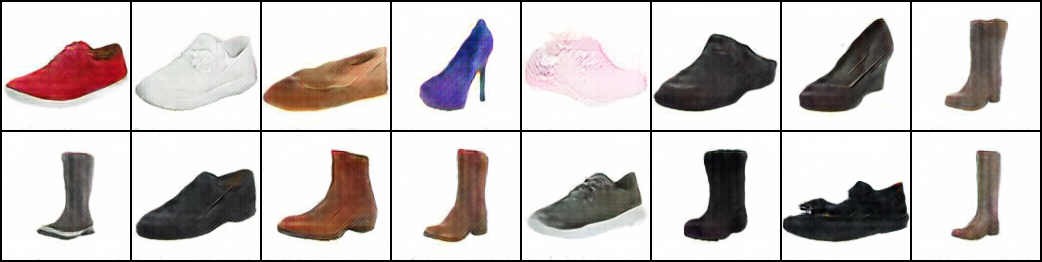}
        \caption{Reconstructed data samples, $G(z^*)$ using a GAN at resolution $128\times128$}
    \end{subfigure}
    \begin{subfigure}{\columnwidth}
        \includegraphics[width=\columnwidth]{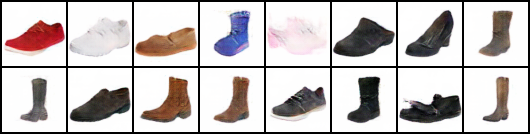}
        \caption{Reconstructed data samples, $G(z^*)$ using a GAN at resolution $64\times64$}
    \end{subfigure}
    \begin{subfigure}{\columnwidth}
        \includegraphics[width=\columnwidth]{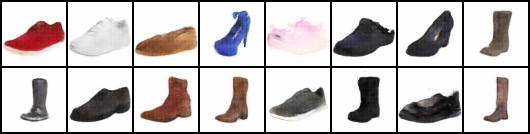} 
        \caption{Reconstructed data samples, $G(z^*)$ using a WGAN at resolution $64\times64$}
    \end{subfigure}
    \caption{\textbf{Reconstruction of Shoes.} By comparing reconstructions, particularly of the blue shoe, we see that the higher resolution model (b) is able to capture some structural details, specifically the shoe's heel, that the lower resolution model (c) does not. Further, the WGAN (d) is able to capture additional detail, including the blue shoe's strap. These results demonstrate how inversion may be a useful tool for comparing which features of a dataset each model is able to capture.}
    \label{fig:shoes_rec}
\end{figure}

In Figure \ref{fig:shoes_rec} we compare shoe reconstructions using a DCGAN trained on low and high resolution images. By comparing reconstructions in Figures \ref{fig:shoes_rec} (b) and (c) (particularly the blue shoe on the top row) we see that the lower resolution model has failed to capture some structural details, while the higher resolution model has not. This suggests that the model trained on higher resolution images is able to capture more structural details than the model trained on lower resolution images. Using our inversion technique to make comparisons between models is just one example of how inversion may also be used to ``trouble shoot'' and identify which features of a dataset our models are not capturing.

Additionally, we may observe that while the GAN trained on higher resolution images preserves more structure than the GAN trained on lower resolution images, it still misses certain details. For example the reconstructed red shoes do not have laces (top left Figure \ref{fig:shoes_rec}(b,c)) laces. This suggests that the representation is not able to distinguish shoes with laces from those without. This may be important when designing representations for image retrieval, where a retrieval system using this representation may be able to consistently retrieve red shoes, but less consistently retrieve red shoes with laces. This is another illustration of how a good inversion technique may be used to better understand what representation is learned by a GAN.

Figure \ref{fig:shoes_rec}(d) shows reconstructions using a WGAN trained on low resolution images. We see that the WGAN is better able to model the blue shoe, and some ability to model the ankle strap, compared to the GAN trained on higher resolution images. It is however, difficult to asses from reconstructions, which model best represents the data. In the next Section (\ref{sec:troubleshooting}), we show how our inversion approach may be used to quantitatively compare these models, and determine which learns a better (latent) representation for the data.

Finally, we found that while regularisation of the latent space may not always improve reconstruction fidelity, it can be helpful for ensuring that latent encodings, $z^*$, found through inversion, correspond to images, $G(z^*)$ that look more like shoes. Our results in Figure \ref{fig:shoes_rec} were achieved using $\beta=0.01$.

\subsection{CelebA}

\begin{figure}[h]
    \begin{subfigure}{\columnwidth}
        \includegraphics[width=\columnwidth]{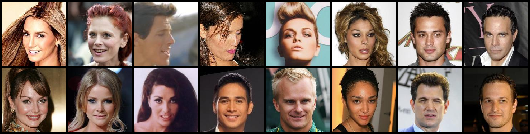}
        \caption{CelebA faces, $x$, from a test set}
    \end{subfigure}
    \begin{subfigure}{\columnwidth}
        \includegraphics[width=\columnwidth]{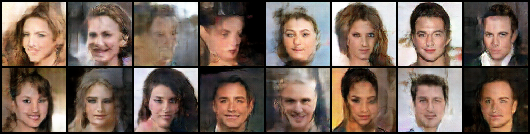}
        \caption{Reconstructed data samples, $G(z^*)$, using a WGAN}
    \end{subfigure}
    \begin{subfigure}{\columnwidth}
        \includegraphics[width=\columnwidth]{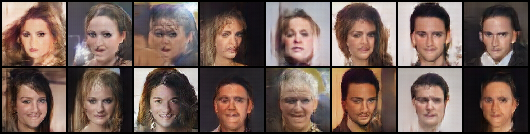}
        \caption{Reconstructed data samples, $G(z^*)$, using a GAN+noise}
    \end{subfigure}
    \begin{subfigure}{\columnwidth}
        \includegraphics[width=\columnwidth]{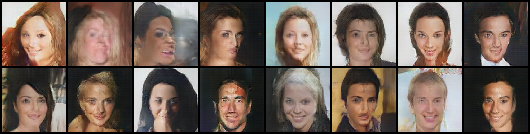}
        \caption{Reconstructed data samples, $G(z^*)$, using a GAN}
    \end{subfigure}
    \caption{\textbf{Reconstruction of celebrity faces}}
    \label{fig:face_rec}
\end{figure}

Figure \ref{fig:face_rec} shows reconstructions using three different GAN models. Training GANs can be very challenging, and so various modifications may be made to their training to make them easier to train. Two examples of modifications are (1) adding corruption to the data samples during training \cite{arjovsky2017towards} and (2) a reformulation of the cost function to use the Wasserstein distance. While these techniques are known to make training more stable, and perhaps also prevent other pathologies found in GANs for example, mode collapse, we are interested to compare the (latent) representations learned by these models.

The most faithful reconstructions appear to be those from the WGAN Figure \ref{fig:face_rec} (b). This will be confirmed quantitatively in the next section. By observing reconstruction results across all models in Figure \ref{fig:face_rec}, it is apparent that all three models fail to capture a particular mode of the data; all three models fail to represent profile views of faces.




\section{Quantitatively Comparing Models}

\label{sec:troubleshooting}

Failing to represent a mode in the data is commonly referred to a ``mode dropping'', and is just one of three common problems exhibited by GANs. For completeness, common problems exhibited by trained GANs include the following: (1) mode collapse, this is where similar image samples are synthesised for different inputs, (2) mode dropping (more precisely), this is where the GAN only captures certain regions of high density in the data generating distribution and (3) training sample memorisation, this is where the GAN memorises and reproduces samples seen in the training data. If a model exhibits these symptoms, we say that it has over-fit, however these symptoms are often difficult to detect.



\begin{table}[h!]
    \centering
    \caption{\textbf{Comparing Models Using Our Inversion Approach} MSE is reported across all test samples for each model trained with each dataset. A smaller MSE suggests that the model is better able to represent test data samples.}
    \label{tab:compare}
    \begin{tabular}{c c c c}
        \textbf{Model}  & CelebA & Shoes & Omniglot\\ \toprule 
        GAN \cite{radford2015unsupervised} & 0.118 & 0.059 & 0.588\\
        GAN$+$noise \cite{arjovsky2017towards} & 0.109 & 0.029 & 0.305\\
        WGAN \cite{arjovsky2017wasserstein} & 0.042 &  0.020 & 0.082\\ \hdashline[1pt/5pt]
        High Res. & -  & 0.016 & - \\ \bottomrule
    \end{tabular}
\end{table}

If a GAN is trained well and exhibits none of the above three problems, it should be possible to preform inversion to find suitable representations for most test samples using our technique.

However, if a GAN does exhibit any of the three problems listed above, inversion becomes challenging, since certain regions of high density in the data generating distribution may not be represented by the GAN. Thus, we may compare GAN models, by evaluating reconstruction error using our proposed inversion process. A high reconstruction error, in this case mean squared error (MSE), suggests that a model has possibly over-fit, and is not able to represent data samples well. By comparing MSE between models, we can compare the extent to which one model has over-fit compared to another. 

Table \ref{tab:compare} shoes how our inversion approach may be used to quantitatively compare $3$ models ($4$ in the case of the shoes dataset) across three datasets, CelebA, shoes and Omniglot. The Table shows mean squared reconstruction error on a large \footnote{CelebA:100 samples, Shoes and Omniglot:500 samples} batch of test samples.

From Table \ref{tab:compare} we may observe the following:

\subsubsection{CelebA}
The (latent) representation learned by the WGAN generalises to test samples, better than either the GAN or the GAN trained with noise. Results also suggests that training a GAN with noise helps to prevent over-fitting. These conclusions are consistent with both empirical and theoretical results found in previous work \cite{arjovsky2017wasserstein, arjovsky2017towards}, suggesting that this approach for quantitatively comparing models is valid.

\subsubsection{Shoes}
Using inversion to quantify the quality of a representation allows us to make fine grain comparisons between models. We see that training a model using higher resolution images reduces reconstruction error by almost a factor of two, in the case of the GAN$+$noise, compared to a similar model trained at a lower resolution. We know from earlier observations (Figure \ref{rec:shoes_rec}), that this is because the model trained at higher resolution, captures finer grain details, that the model trained on lower resolution images.

Comparing models using our proposed inversion approach, in addition to classifier based measures \cite{salimans2016improved}, helps to detect fine grain differences between models, that may not be detected using classification based measures alone. A ``good'' discriminative model learns many features that help to make decisions about which class an object belongs to. However, any information in an image that does not aid classification is likely to be ignored (e.g. when classifying cars and trucks, the colour of a car is not important \footnote{Bottom left of Figure 10 \cite{mahendran2015understanding}, shows that the 1st layer of a discriminatively trained CNN ignores the colours of the input image.}). Yet, we may want to compare representations that encode information that a classifier ignores (e.g. colour of the car). For this reason, using only a classification based measure \cite{salimans2016improved} to compare representations learned by different models may not be enough, or may require very fine grain classifiers to detect differences.



\subsubsection{Omniglot}
From Figure \ref{fig:omni_rec}, it was clear that both models trained on the Omniglot dataset had over-fit, but not to the same extent. Here, we are able to quantify the degree to which each model has over-fit. We see that the WGAN has over-fit to a lesser extent compared to the GAN trained with noise, since the WGAN has a smaller MSE. Quantifying over-fitting can be useful when developing new architectures, and training scheme, to objective compare models.

In this section we have demonstrated how our inversion approach may be used to quantitatively compare representations learned by GANs. We intend this approach to provide a useful, quantitative approach for evaluating and developing new GAN models and architectures for representation learning.

Finally, we emphasise that while there are other techniques that provide inversion, ours is the only one that is \textbf{both} (a) immune to over-fitting, in other words we do not train an encoder network that may itself over-fit, and (b) can be applied to any pre-trained GAN model provided that the computational graph is available.







\section{Conclusion}

The generator of a GAN learns the mapping $G: Z \rightarrow X$. It has been shown that $z$ values that are close in $Z$-space produce images that are visually similar in image space, $X$ \cite{radford2015unsupervised}. We propose an approach to map data, $x$ samples back to their latent representation, $z^*$ (Section \ref{sec:method}).

For a generative model, in this case a GAN, that is trained well and given target image, $x$, we should be able to find a representation, $z^*$, that when passed through the generator, produces an image, $G(z^*)$, that is similar to the target image. However, it is often the case that GANs are difficult to train, and there only exists a latent representation, $z^*$, that captures some of the features in the target image. When $z^*$ only captures some of the features, this results in, $G(z^*)$, being a partial reconstruction, with certain features of the image missing. Thus, our inversion technique provides a tool, to provide qualitative information about what features are captured by in the (latent) representation of a GAN. We showed several visual examples of this in Section \ref{sec:results}.

Often, we want to compare models quantitatively. In addition to providing a qualitative way to compare models, we show how we may use mean squared reconstruction error between a target image, $x$ and $G(z^*)$, to quantitatively compare models. In our experiments, in Section \ref{sec:troubleshooting}, we use our inversion approach to quantitatively compare $3$ models trained on $3$ datasets. Our quantitative results support claims from previous work that suggests, that certain modified GANs are less likely to over-fit.

We expect that our proposed inversion approach may be used as a tool to asses and compare various proposed modifications to generative models, and aid the development of new generative approaches to representation learning.






\section*{Acknowledgements}
We like to acknowledge the Engineering and Physical Sciences Research Council for funding through a Doctoral Training studentship.

\bibliographystyle{abbrv}
\bibliography{bib}


\end{document}